\documentclass[conference]{IEEEtran}
\usepackage[T1]{fontenc}
\usepackage{graphicx}
\usepackage{subcaption}
\usepackage{color}
\usepackage{textcomp}
\usepackage{csquotes}
\usepackage{hyperref}

\usepackage{caption} 
\usepackage{tabularx}
\usepackage{float}
\usepackage{array}  
\usepackage{booktabs} 
\usepackage{multirow}
\usepackage{makecell}
\usepackage{arydshln} 
\usepackage{todonotes}
\usepackage{eurosym}
\usepackage{balance}

\usepackage{siunitx}
\usepackage{cite}
\begin{document}

\title{Touched by ChatGPT: Using an LLM to Drive Affective Tactile Interaction}

\author{\IEEEauthorblockN{Qiaoqiao Ren}
\IEEEauthorblockA{\textit{Faculty of Engineering and Architecture} \\
\textit{IDLab-AIRO, Ghent University -- imec}\\
Qiaoqiao.Ren@ugent.be}
\and
\IEEEauthorblockN{Tony Belpaeme}
\IEEEauthorblockA{\textit{Faculty of Engineering and Architecture} \\
\textit{IDLab-AIRO, Ghent University -- imec}\\
Tony.Belpaeme@ugent.be}

}



\markboth{Journal of \LaTeX\ Class Files,~Vol.~14, No.~8, August~2021}%
{Shell \MakeLowercase{\textit{et al.}}: A Sample Article Using IEEEtran.cls for IEEE Journals}


\maketitle

\begin{abstract}


Touch is a fundamental aspect of emotion-rich communication, playing a vital role in human interaction and offering significant potential in human-robot interaction. Previous research has demonstrated that a sparse representation of human touch can effectively convey social tactile signals. However, advances in human-robot tactile interaction remain limited, as many humanoid robots possess simplistic capabilities, such as only opening and closing their hands, restricting nuanced tactile expressions. In this study, we explore how a robot can use sparse representations of tactile vibrations to convey emotions to a person. To achieve this, we developed a wearable sleeve integrated with a 5x5 grid of vibration motors, enabling the robot to communicate diverse tactile emotions and gestures. Using chain prompts within a Large Language Model (LLM), we generated distinct 10-second vibration patterns corresponding to 10 emotions (e.g., happiness, sadness, fear) and 6 touch gestures (e.g., pat, rub, tap). Participants ($N$ = 32) then rated each vibration stimulus based on perceived valence and arousal. People are accurate at recognising intended emotions, a result which aligns with earlier findings. These results highlight the LLM's ability to generate emotional haptic data and effectively convey emotions through tactile signals. By translating complex emotional and tactile expressions into vibratory patterns, this research demonstrates how LLMs can enhance physical interaction between humans and robots. 

\end{abstract}


\section{Introduction}

Emotion is central to human communication, influencing interactions, decision-making, and relationships \cite{assunccao2022overview}. One promising approach is using haptic feedback, particularly vibration, to convey emotional states. Vibration is an intuitive and non-verbal channel that can convey a range of emotional cues \cite{mazzoni2015does}, making it a valuable tool for enriching communication in digital systems where the use of traditional sensory modalities, such as vision and sound, may be limited \cite{loomis2007functional}.

Haptic devices have long been explored for their role in augmenting emotional communication. Early studies demonstrated the ability of vibrations to convey basic emotions, such as happiness or sadness, through simple intensity modulation \cite{eid2015affective}. Researchers developed wearable haptic devices to convey emotional states, finding moderate success in user decoding, particularly for high-arousal emotions \cite{olugbade2023touch}. Similarly, Salminen \cite{salminen2015emotional} used force feedback and skin stretch to evoke emotional responses, demonstrating that tactile communication could enhance immersion and empathy in digital environments. However, these approaches often rely on manually designed patterns, which can limit their scalability and effectiveness in representing complex emotions.

Russell's circumplex model of emotions \cite{russell1980circumplex}, which organizes emotions along valence (pleasantness) and arousal (intensity) dimensions, provides a useful framework for designing haptic feedback systems. Studies have used this model to map emotions to haptic parameters such as vibration intensity, rhythm, and duration, enabling more structured approaches to haptic communication \cite{teyssier2020conveying}. However, these efforts often focus on a limited set of emotions and fail to account for the intricacies of emotional expressions or interpersonal variability in emotion decoding.

\begin{figure*}
\centering
\begin{subfigure}{0.3\textwidth}
    \centering
    \includegraphics[height=4.2cm]{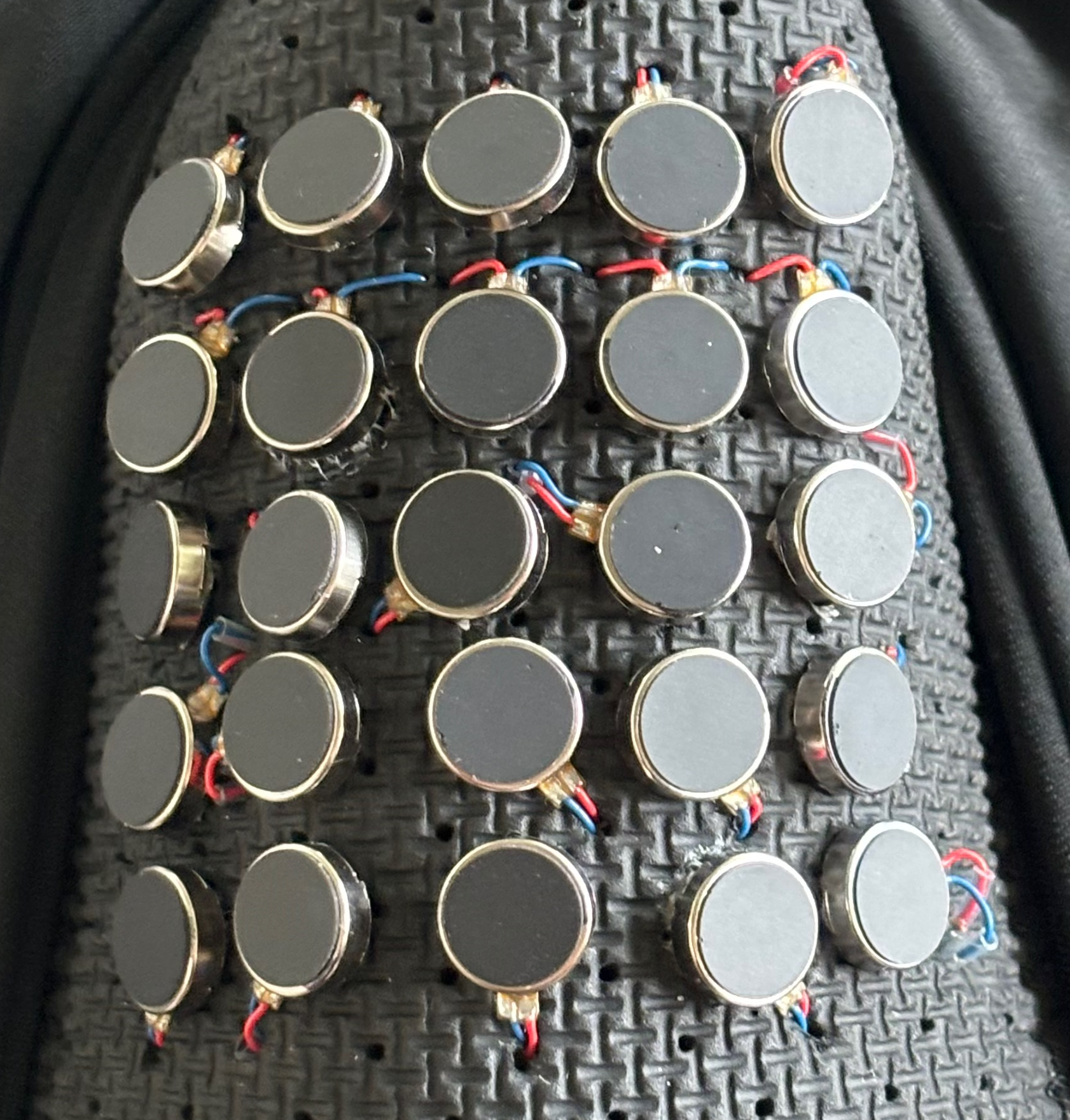}
\caption{Vibration motors distribution.}
    \label{fig:motors1}
\end{subfigure}
\begin{subfigure}{0.3\textwidth}
    \centering
    \includegraphics[height=4.2cm]{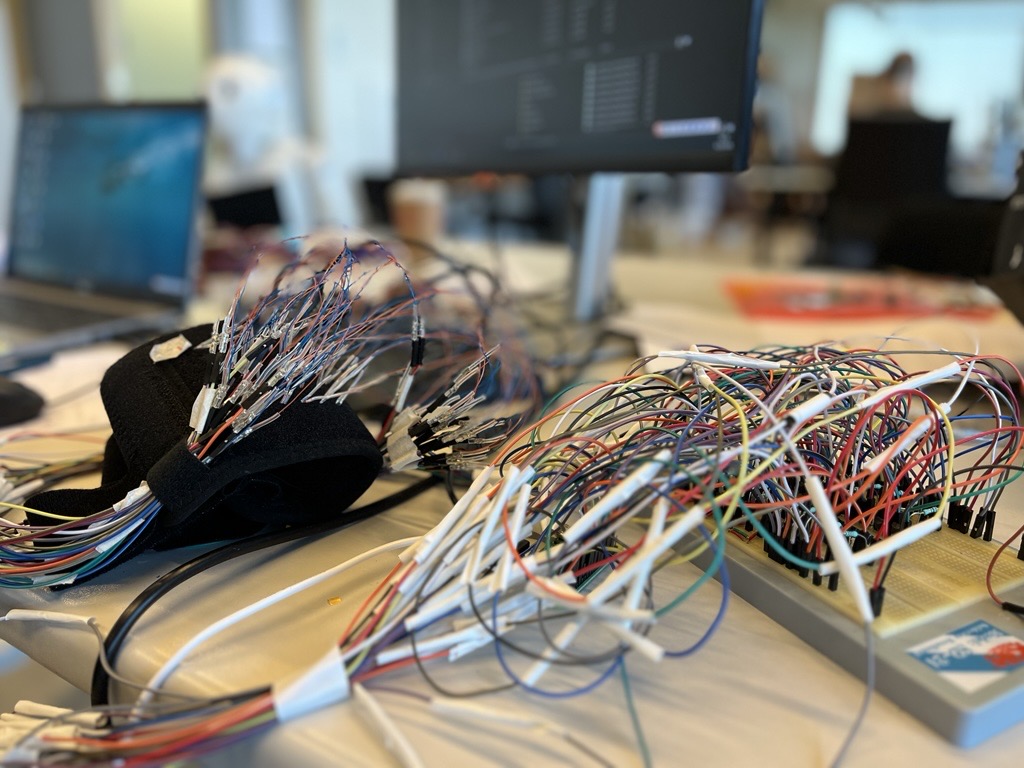}
    \caption{Vibration sleeve.}
    \label{fig:whole_device}
\end{subfigure}
\begin{subfigure}{0.38\textwidth}
    \centering
    \includegraphics[height=4.2cm]{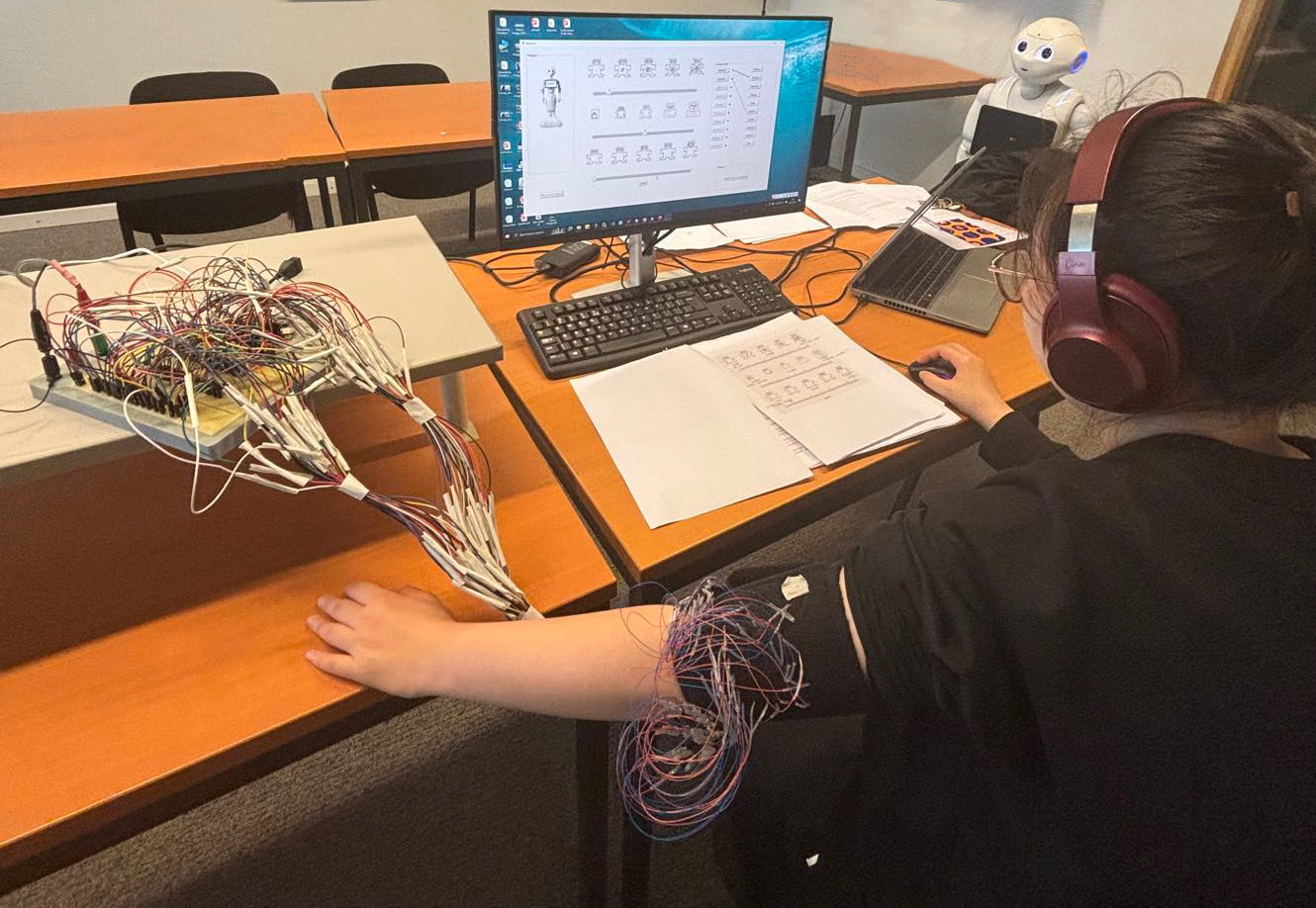}
    \caption{Convey emotions to the participants.}
    \label{fig:interaction}
\end{subfigure}
\caption{Experimental setup.}
\label{fig:experimentalsetup}
\end{figure*}

The application of AI in emotion communication has been explored in visual and auditory modalities \cite{shoumy2020multimodal}, but its integration into haptic systems remains relatively underdeveloped \cite{giri2021application}. Traditional approaches for generating tactile patterns often rely on non-AI methods, such as heuristic-based systems \cite{saket2013designing, mathew2005vsmileys}. These methods have primarily focused on understanding how features --such as vibration frequency, amplitude, and duration-- influence participants' perceptions of arousal and valence \cite{seifi2013first}. While insightful, these approaches face significant limitations, including hardware constraints, differences in the contact areas of vibration motors, variability in motor properties, and challenges in standardizing tactile stimuli across participants \cite{schneider2017haptic}. Therefore, it is still a significant challenge to reflect complex emotional expressions accurately using tactile communication. Previous studies have shown that even in human-human communication, decoding emotional signals can be inherently difficult. For example, the accuracy of emotion decoding between individuals in close relationships, such as couples, often falls below 50\% for a limited set of emotions \cite{thompson2011effect}. These findings underscore the need for more expressive and reliable haptic systems. Moreover, previous studies have predominantly focused on collecting affective data directly from participants, relying on manually designed patterns that may not capture the full range of emotional nuances. Recent developments in LLMs have introduced new possibilities in this area. LLMs, with their ability to analyse large datasets, offer a powerful tool for understanding the features of emotional expressions and translating them into tactile feedback \cite{yu2024octopi}. For instance, Yu \textit{et al.} \cite{yu2024octopi} demonstrated the potential of tactile-language models in object property reasoning, bridging the gap between tactile perception and linguistic representation. 


Our study uses a 5x5 grid of vibration motors embedded in a wearable device, designed to convey affective touch. We use an LLM to control the device, a notable and unique contribution. The LLM is first prompted to analyse the tactile features of emotions such as happiness, sadness, fear, and anger, as well as touch gestures like patting and rubbing. Next, the LLM generates corresponding vibration patterns based on the analysis. These patterns are mapped onto the 5x5 grid motors to produce varying intensities, rhythms, and spatial vibration distributions. By integrating LLM-driven analysis into the design process, we aim to create more expressive and dynamic haptic feedback systems capable of conveying various emotions. Our primary research question explores whether a LLM can generate tactile data that effectively conveys emotions to humans.


\section{Materials \& Methods}

To validate the ability of an LLM to generate tactile signals, we collect data on how people interpret these stimuli. Participants are tasked with decoding 10 emotions (anger, fear, disgust, happiness, surprise, sadness, confusion, comfort, calm, and attention) and 6 gestures (hold, pat, tickle, rub, tap, and poke). Before data collection, participants were provided with definitions of all emotions (to ensure consistent interpretations for non-native English-speaking participants) and touch gestures to ensure they understood and agreed with the given definitions. Participants are asked to rate the arousal and valence on a scale from 1 to 10 for each stimulus generated by the LLM. Subjective ratings of arousal and valence are given on a 10-point scale, with lower scores indicating lower arousal and less pleasant stimuli, and higher scores reflecting higher arousal and more pleasant stimuli. Afterwards, they are asked to classify the stimulus as expressing a specific emotion or gesture. The following sections provide details on the equipment, the study design, and the data acquisition.

\label{sec2:Materials}

\subsection{Equipment}



\subsubsection{Vibration sleeve}

The hardware for the vibration device consists of a 5x5 grid of vibration motors embedded in an upper arm sleeve, powered by a 3.3V power supply. Each motor is controlled by a single Raspberry Pi using BC557B PNP transistors, which act as switches to regulate current flow to the motors, as shown in Fig.~\ref{fig:motors1} and Fig.~\ref{fig:whole_device}. The system uses pulse-width modulation (PWM) for precise control over the intensity of each motor's vibration, which allows for modulation of the vibration intensity by rapidly switching the transistors on and off at varying duty cycles. By adjusting the duty cycle of the PWM signal, the perceived intensity of each motor’s vibration can be finely controlled, allowing for a dynamic range of vibration patterns. 

\subsubsection{Prompts}

\begin{figure}[h]
\centering
    \centering
    \includegraphics[height=3cm]{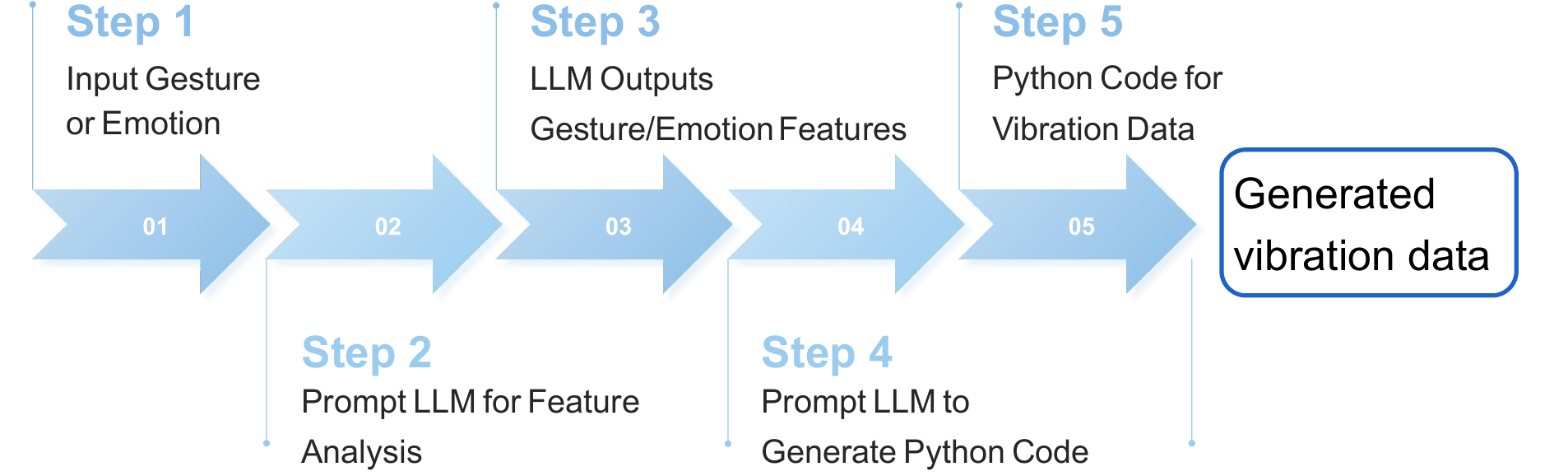}
\caption{Flow chart of LLM generated vibration data.}
    \label{fig:Flow_chart}
\end{figure}

\begin{figure*}
    \centering
        \includegraphics [width=\textwidth]{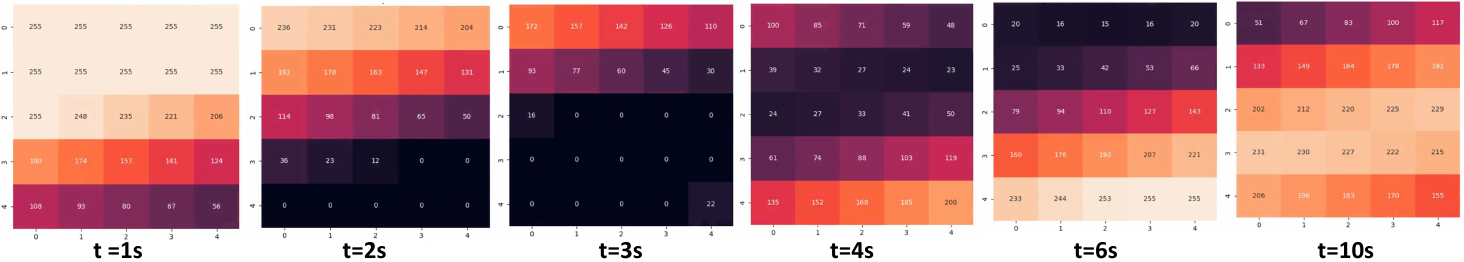}
    \caption{Illustration of frames of LLM generated \textit{rub} gesture.}
    \label{fig:visualization}
\end{figure*}

This approach uses a chain of prompts to guide an LLM (\textit{gpt-4o}) to generate vibration data for a 5x5 grid of vibration motors. 
Inspired by previous work where 10-second data was collected from humans and transformed into vibration patterns to convey emotions \cite{ju2021haptic}, we asked the LLM to generate 10 seconds of vibration data. The vibration data generation flow chart is shown in Fig.~\ref{fig:Flow_chart}. The process begins by prompting the LLM to analyze the features of a specific gesture or emotion (e.g., happiness, sadness, rub, tap). The LLM is asked how such an emotion or gesture would manifest in terms of pressure on the grid, focusing on how the vibration magnitude changes over time and across the spatial arrangement of vibration motors. The prompt also directs the model to account for smooth transitions in pressure values, reflecting the natural progression of touch or emotional cues. This ensures a natural onset and decay of the motors.

Once the analysis is complete, a second prompt uses the insights from the first to generate Python code. This code simulates the vibration data by creating a data frame, capturing the evolving vibration patterns over time. The Python code ensures that the data is in a format that can be used directly to control the 5x5 grid of vibration motors. Key requirements like smooth pressure transitions and sustained changes over multiple time steps are emphasized to ensure the data is suitable for producing realistic haptic feedback. The output of this process is a CSV file for 10 seconds of vibration data. The motor distributions and visualization of LLM-generated data are shown in Fig.~\ref{fig:motors1} and Fig.~\ref{fig:visualization}. The detailed prompts, code and visualization can be found in our GitHub repository\footnote{\url{https://github.com/TouchedByLLM/LLM_generate_data}}.

\section{Experiment setup}

To explore emotion decoding via touch-based expression and reveal if and how people can decode the emotions and gestures generated by the LLM, we set up a data collection experiment (Fig.~\ref{fig:interaction}). We recruited 32 participants on a university campus, 17 identified as female, 15 as male, average age $27.5\pm 3.6$ years. The data collection and study complied with the ethics regulations of \emph{ANONYMOUS} and participants gave informed consent before data collection. The collection was split into two sessions: in the first session they  \textit{decode emotions conveyed by the robot}, in the second session they \textit{decode tactile gestures conveyed from the robot}.

\begin{enumerate}

    \item \textit{Pre-session:} Participants were fitted with the vibration device and their perception threshold was set. 
       
    \item \textit{Decoding Emotions from the Robot:} Participants were asked to interpret emotions from 10 stimuli delivered by the robot through the wearable device. For each stimulus, they rated arousal and valence and identified the specific emotion. Each stimulus lasted 10 seconds, with time provided between stimuli for participants to provide their responses. The next stimulus was offered once participants indicated they were ready. 
    
    \item \textit{Decoding Tactile Gestures from the Robot:} In this session, participants decoded touch gestures from 6 stimuli delivered by the robot through the wearable device. For each stimulus, they identified the corresponding gesture and could replay it as needed until they were sure of their answer. Each stimulus lasted 10 seconds, with preparation time given between stimuli. As before, the next stimulus was offered when participants indicated they were ready.

\end{enumerate}

\section{Results and discussion}

\subsection{Emotion decoding}

\begin{figure}[b]
\centering
    \centering
    \includegraphics[height=5cm]{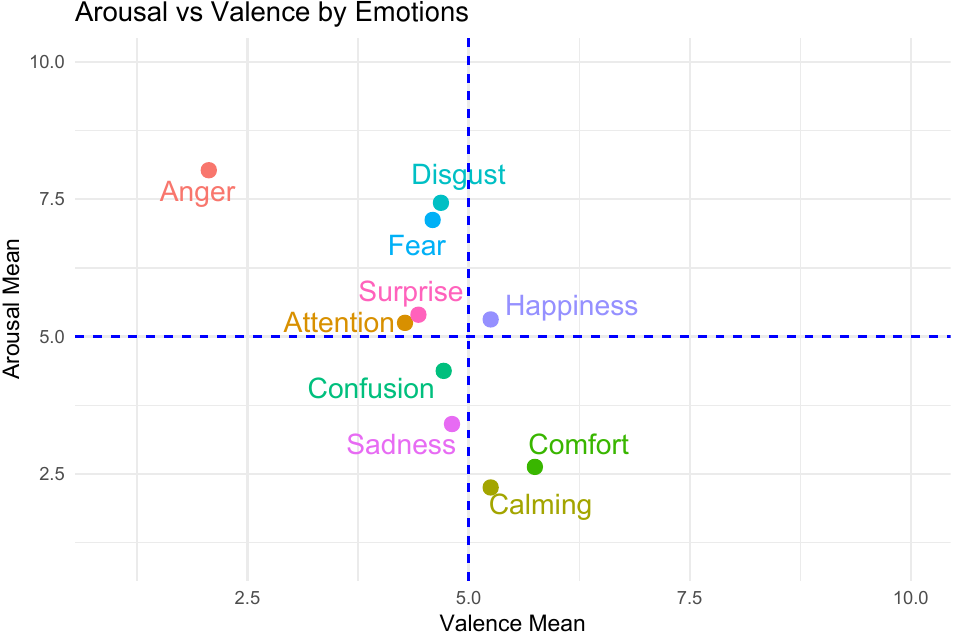}
\caption{Arousal and valence rating distribution of emotions.}
    \label{fig:emotion_va}
\end{figure}

We averaged participants' arousal and valence ratings for different emotional stimuli, which are shown in Table.~\ref{tab:emotion_va}. 
Figure.~\ref{fig:emotion_va} illustrates the distribution of the 10 emotions within the emotion complex model. Participants on average placed \textit{happiness} in the high-arousal, positive-valence quadrant. \textit{Disgust}, \textit{fear}, \textit{surprise} and \textit{anger} were placed in the high-arousal, negative-valence quadrant. \textit{Confusion} and \textit{sadness} were assigned to the low-arousal, negative-valence quadrant, while \textit{comfort} and \textit{calming} were grouped in the low-arousal, positive-valence quadrant. This classification is consistent with the emotion categorisation described in Russell's complex model and previous researches.\cite{noordewier2013valence, bradley2001emotion}.

\begin{table*}[t]\footnotesize
\setlength{\abovecaptionskip}{0.0cm}   
	\setlength{\belowcaptionskip}{0.0cm}  
	\renewcommand\tabcolsep{2.0pt} 
	\centering
	\caption{Arousal and valence of different emotions and decoding accuracy (\%)}
	\begin{tabular}
	{
	p{1.5cm}<{\centering} 
 p{1cm}<{\centering} 
	 p{1.5cm}<{\centering} 
  p{1.5cm}<{\centering}
	p{1cm}<{\centering}
 	p{1.5cm}<{\centering}
        p{1cm}<{\centering}
 	p{1.5cm}<{\centering}
 	p{1.5cm}<{\centering}
 	p{1.5cm}<{\centering}
 	p{1cm}<{\centering}
	} 
\hline

     {Emotions} & {Happiness} & {Surprise}  & {Fear} & {Disgust} & {Anger} & {Comfort}  & {Attention} & {Calming} & {Confusion} & {Sadness} \\

\hline
   {Arousal} & {$5.3\pm2.0$} & {$5.3\pm2.2$} & {$7.1\pm1.9$} & {$7.4\pm1.7$} & {$8.0\pm1.5$} & {$2.6\pm1.9$} & {$5.3\pm1.9$} & {$2.3\pm1.9$} & {$4.4\pm2.1$} & {$3.4\pm2.1$}\\
   {Valence} & {$5.3\pm2.3$} & {$4.4\pm1.9$} & {$4.6\pm2.8$} & {$4.7\pm2.5$} & {$2.1\pm1.4$} & {$5.8\pm1.5$} & {$4.3\pm1.8$} & {$5.3\pm2.0$} & {$4.7\pm2.3$} & {$4.8\pm2.2$}\\
    {Accuracy} & {$21.9$} & {$\textbf{25.0}$} & {$\textbf{31.2}$} & {$21.9$} & {$\textbf{68.8}$} & {$\textbf{25.0}$} & {$\textbf{31.2}$} & {$\textbf{37.5}$} & {$12.5$} & {$28.1$}\\
    {Sig.(p)} & {$0.06$} & {$0.03$} & {$<0.01$} & {$0.06$} & {$<0.01$} & {$0.03$} & {$<0.01$} & {$<0.01$} & {$0.42$} & {$0.28$}\\

     \hline

    \end{tabular}
\label{tab:emotion_va}
\end{table*}

\begin{figure*}
\centering
\begin{subfigure}{0.6\textwidth}
    \centering
    \includegraphics[height=4.5cm]{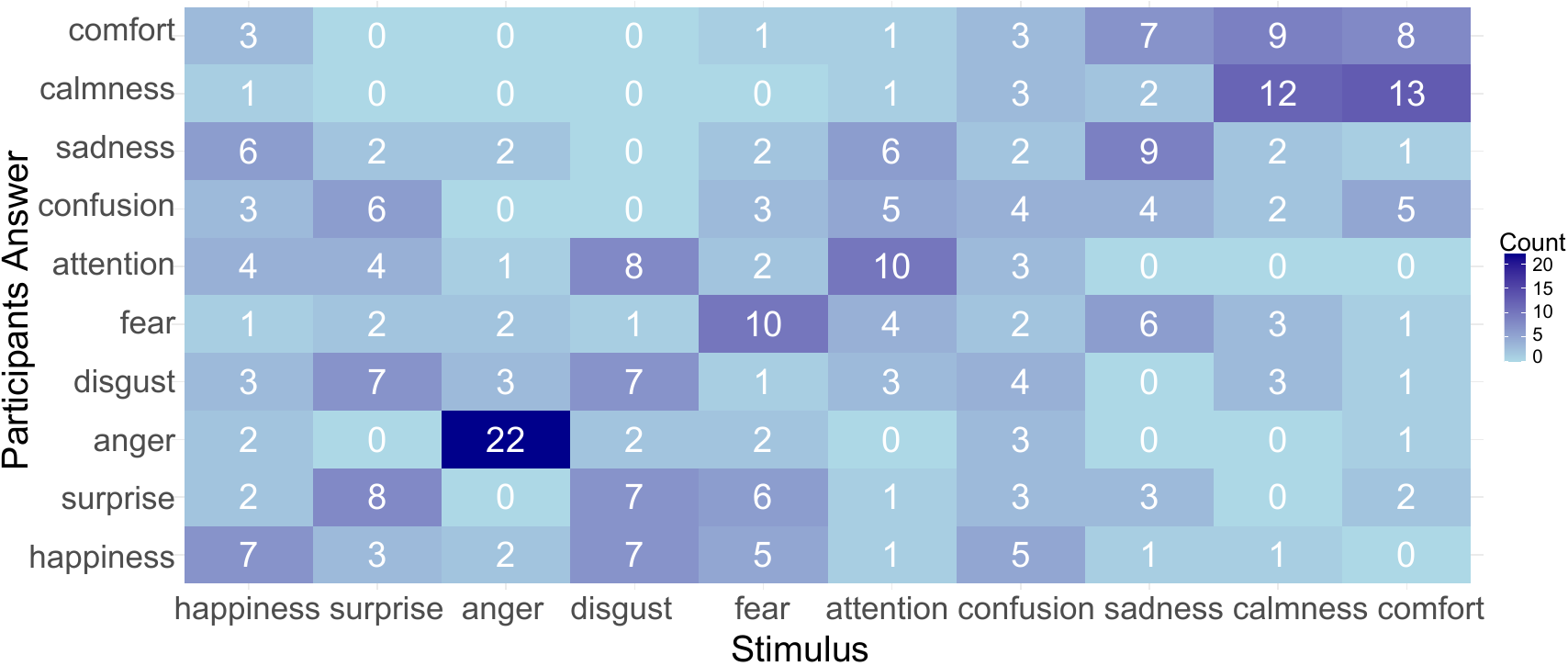}
\caption{Confusion matrix for emotions}
    \label{fig:emotion_confusion}
\end{subfigure}
\begin{subfigure}{0.35\textwidth}
    \centering
    \includegraphics[height=4.5cm]{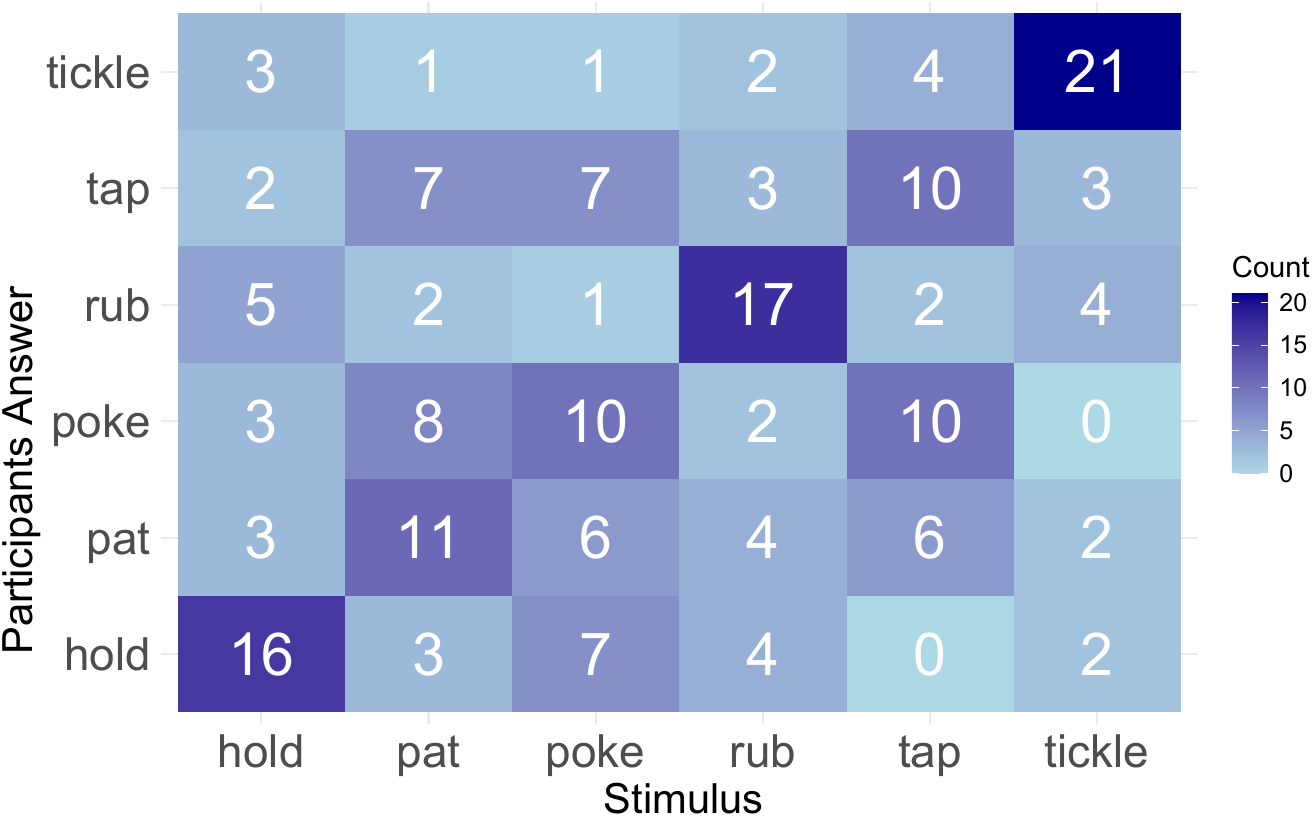}
    \caption{Confusion matrix for touch gestures}
    \label{fig:gesture_confusion}
\end{subfigure}
\caption{Confusion matrix.}
\label{fig:confusion_matrix}
\end{figure*}





The classification results in Fig.~\ref{fig:emotion_confusion} showed that 22 out of 32 participants were correctly classified \textit{anger}, making it the most accurately decoded emotion. This was followed by \textit{comfort}, which 13 participants decoded correctly, and \textit{calming}, decoded correctly by 12 participants. The overall decoding accuracy across the 10 emotions is 30.3\%. Previous research \cite{nummenmaa2014bodily} compared participant responses to the probability of randomly guessing correctly (the ``chance level''). In our experiment, the chance level for emotions is 10\%. Previous research \cite{thompson2011effect, hertenstein2006touch} explored emotion decoding among couples and strangers, and the results we obtained in our research are similar to the results they presented for strangers (37.5\%). If we set the 37.5\% as a baseline, we used a one-sample t-test to determine whether the participant's decoding accuracy is significantly higher than chance level; the results showed that there is no significant difference between human-to-human emotion decoding among strangers in previous research and our human-robot decoding ($t(31) = -2.99, p = 0.99$). And Table.~\ref{tab:emotion_va} presents the decoding accuracy for each emotion, along with its statistical significance level, in comparison to the chance level. In addition, compared to a chance level of 10\%, the participants' decoding results (30.3\%) are significantly above chance ($t(31) = 7.89, p<0.001$). Therefore, we can conclude that the participants succeeded in decoding the emotional data generated by the LLM, and the LLM has the ability to generate affective touch.






\begin{table}[h]
\footnotesize
\setlength{\abovecaptionskip}{0.0cm}   
	\setlength{\belowcaptionskip}{0.0cm}  
	\renewcommand\tabcolsep{2.0pt} 
	\centering
	\caption{Decoding accuracy(\%)}
	\begin{tabular}
	{
	p{1.5cm}<{\centering} 
 p{1cm}<{\centering} 
	 p{1.5cm}<{\centering} 
  p{1cm}<{\centering}
	p{1cm}<{\centering}
 	p{1cm}<{\centering}
        p{1cm}<{\centering}
	} 
\hline
      
     {Gestures} & {Hold} & {Pat}  & {Poke} & {Rub} & {Tap} & {Tickle}  \\
     
\hline
   {Accuracy} & {$53.1$} & {$34.4$} & {$31.2$} & {$53.1$} & {$31.2$} & {$65.6$} \\
    {Sig.(p)} & {$<0.01$} & {$0.02$} & {$0.045$} & {$0.01$} & {$0.045$} & {$<0.01$} \\

     \hline

    \end{tabular}
\label{tab:gesture}
\end{table}

\subsection{Gesture decoding}

As shown in the Table.~\ref{tab:gesture} and Fig.~\ref{fig:gesture_confusion}, the gesture that participants found easiest to decode was \textit{tickle}, followed by \textit{rub}. In contrast, \textit{poke}, \textit{pat} and \textit{tap} were more frequently misclassified and often confused with each other. One possible explanation for this confusion is the limited size of the designed sleeve's touch-sensitive area. Although the 25-motor setup aims to facilitate relative touch perception, the small touch area (6cm by 6cm) on the upper arm may hinder participants from distinguishing significant differences between certain gestures. For instance, while \textit{poke} involves a smaller contact area compared to \textit{pat}, their rhythmic patterns can be similar, as is the case with \textit{tap}. This overlap in tactile cues may lead to confusion and reduce the fidelity of gesture recognition when conveying gestures to participants. The participants' touch gesture decoding accuracy is significantly higher than the chance level of 16.7\% ($t(31) = 7.67, p<0.001$). Overall, participants decoded all touch gestures significantly higher than the chance level, while they could only decode six emotions significantly higher than the chance level.

\subsection{Conclusion and future work}

This study highlights the potential of using a wearable sleeve with a 5x5 grid of vibration motors, combined with LLM-generated vibration patterns, to effectively convey emotions and touch gestures. Participants successfully decoded 10 emotions and 6 gestures with accuracy levels that matched prior research in human-human tactile interaction and were significantly higher than chance. Notably, the results demonstrate that LLMs, despite never having been trained on tactile interaction experiences, can generate affective touch data purely based on textual information and descriptions. However, some emotions, such as \textit{confusion}, did not perform significantly higher than the chance level. This is reasonable, as certain emotions are inherently more challenging to convey purely through touch; most participants also mentioned that they feel confusion and disgust are more challenging to decode than other emotions. These emotions often rely on nuanced contextual or introspective cues that are difficult to translate into tactile patterns alone. In addition, some gestures, such as \textit{poke}, \textit{pat}, and \textit{tap}, posed challenges due to the limited tactile area of the sleeve, which may have hindered participants' ability to distinguish between subtle variations in the contact area and rhythm. This suggests opportunities for enhancing the design to improve gesture differentiation. The findings validate the system’s capacity to translate complex emotional and tactile expressions by using LLM into vibratory signals, significantly advancing physical interaction between humans and robots. Future work will aim to refine the tactile interface, enhance gesture distinctiveness, and expand the repertoire of emotional and haptic patterns.

\bibliographystyle{ieeetr}
\bibliography{references}
\balance
\vfill\pagebreak

\end{document}